\documentclass[conference]{IEEEtran}
\IEEEoverridecommandlockouts
\usepackage{cite}
\usepackage{amsmath,amssymb,amsfonts}
\usepackage{algorithmic}
\usepackage{graphicx}
\usepackage{textcomp}
\usepackage{booktabs}
\usepackage{xcolor}
\usepackage{caption}
\usepackage{subcaption}
\usepackage{pifont}
\usepackage{tikz}
\usepackage{hyperref}
\def\BibTeX{{\rm B\kern-.05em{\sc i\kern-.025em b}\kern-.08em
    T\kern-.1667em\lower.7ex\hbox{E}\kern-.125emX}}

\newcommand{\cmark}{\ding{51}}
\newcommand{\xmark}{\ding{55}}
    
\begin{document}

\newcommand\copyrighttext{%
  \footnotesize \textcopyright 2023 IEEE.  Personal use of this material is permitted.  Permission from IEEE must be obtained for all other uses, in any current or future media, including reprinting/republishing this material for advertising or promotional purposes, creating new collective works, for resale or redistribution to servers or lists, or reuse of any copyrighted component of this work in other works.
  DOI: \href{https://doi.org/10.1109/ICDMW60847.2023.00110}{10.1109/ICDMW60847.2023.00110}}
\newcommand\copyrightnotice{%
\begin{tikzpicture}[remember picture,overlay]
\node[anchor=south,yshift=10pt] at (current page.south) {\fbox{\parbox{\dimexpr\textwidth-\fboxsep-\fboxrule\relax}{\copyrighttext}}};
\end{tikzpicture}%
}

\title{ViGEO: an Assessment of Vision GNNs in Earth Observation\\
}

\author{
\IEEEauthorblockN{Luca Colomba}
\IEEEauthorblockA{\textit{Department of Control and Computer Engineering} \\
\textit{Politecnico di Torino}\\
Turin, Italy \\
luca.colomba@polito.it}
\and
\IEEEauthorblockN{Paolo Garza}
\IEEEauthorblockA{\textit{Department of Control and Computer Engineering} \\
\textit{Politecnico di Torino}\\
Turin, Italy \\
paolo.garza@polito.it}
}

\maketitle
\copyrightnotice

\begin{abstract}
Satellite missions and Earth Observation (EO) systems represent fundamental assets for environmental monitoring and the timely identification of catastrophic events, long-term monitoring of both natural resources and human-made assets, such as vegetation, water bodies, forests as well as buildings. Different EO missions enables the collection of information on several spectral bandwidths, such as MODIS, Sentinel-1 and Sentinel-2. Thus, given the recent advances of machine learning, computer vision and the availability of labeled data, researchers demonstrated the feasibility and the precision of land-use monitoring systems and remote sensing image classification through the use of deep neural networks. Such systems may help domain experts and governments in constant environmental monitoring, enabling timely intervention in case of catastrophic events (e.g., forest wildfire in a remote area). Despite the recent advances in the field of computer vision, many works limit their analysis on Convolutional Neural Networks (CNNs) and, more recently, to vision transformers (ViTs). Given the recent successes of Graph Neural Networks (GNNs) on non-graph data, such as time-series and images, we investigate the performances of a recent Vision GNN architecture (ViG) applied to the task of land cover classification. The experimental results show that ViG achieves state-of-the-art performances in multiclass and multilabel classification contexts, surpassing both ViT and ResNet on large-scale benchmarks.
\end{abstract}

\begin{IEEEkeywords}
earth observation, image classification, graph neural networks, deep learning
\end{IEEEkeywords}

\section{Introduction}
Extreme climate events are becoming more and more frequent and endanger both the natural ecosystem and society. The development of physical and machine learning models able to predict in advance such events in conjunction with the development of real-time monitoring systems represent fundamental tasks towards the climate change problem, with the primary goal of limiting as much as possible collateral damages. In this context, the joint use of Earth Observation (EO) systems and satellite imagery with deep learning methodologies enables the creation of large- and global-scale monitoring systems. Resource monitoring from aerial images can be categorized into several different tasks among the research community: image segmentation problem, such as the burned area delineation problem~\cite{double_step_unet} and flood detection~\cite{flood_segmentation}, building and road detection~\cite{road_detection}, air pollution prediction~\cite{air_quality, air_quality_sentinel5} and many others. This work focuses on the land cover (land use) classification task, in which data is categorized into several classes, including vegetation, water reservoir and buildings.

All the aforementioned tasks in the field of EO can be formulated as computer vision problems, extending the input data over the RGB case due to satellite's acquisitions characteristics\footnote{Dozens of channels usually characterize satellite images.} given the different wavelength in which data is collected. Consequently, such field represents an interesting research area in the computer vision domain, introducing more complexity than the regular visible (RGB) case. Since the development of convolutional neural networks (CNNs), multiple studies regarding the adoption of deep learning in EO have been conducted. Few examples are the application of siamese convolutional networks for remote sensing scene classification~\cite{liu2019siamese} and the identification of solar panels through the use of deep CNNs~\cite{malof2017solar}.
More recently, also the remote sensing domain have seen the rise of the adoption of transformer-based architectures~\cite{attention_transformer} to solve the mentioned tasks, demonstrating better performances compared to CNN-based architectures~\cite{cambrin2022vision} in case of great data availability. This is caused by the absence of intrinsic inductive biases which characterize the CNN architecture~\cite{vit}.

In parallel, researchers started investigating the application of graph neural networks on non-graph-structured data, such as regular and irregular time-series data~\cite{gnn_matrix_construction}. In such cases, a simple graph structure is imposed a-priori (e.g., based on distances)~\cite{gnn_timeseries} or is automatically infererred by the neural network~\cite{raindrop}. Few works investigated the application of GNNs to the vision domain for different tasks, mainly related to point clouds~\cite{gnn_cv_survey}, with the Vision GNN architecture~\cite{vig} (ViG) being the most successful architecture in image classification, achieving higher performances in the image classification task compared to the ViT architecture~\cite{vit}.

In this context, we explore the applicability of ViG architecture to the EO domain on a large-scale and multilabel land cover benchmark dataset, namely BigEarthNet~\cite{bigearthnet1, bigearthnet2}, and two other smaller datasets, RESISC45~\cite{resisc45} and PatterNet~\cite{patternnet}. More specifically, we adapted the original formulation of ViG to tailor it to the multispectral imaging context, enabling its use to datasets characterized by smaller input resolutions in terms of pixels. To the best of our knowledge, this is the first application of a Vision GNN model to an extensive benchmark dataset with multilabel classification in the Earth Observation domain.

The contribution of our work can be summarized as follows.
\begin{itemize}
    \item A revised ViG architecture, suitable to process multispectral data and input images with smaller resolution compared to ImageNet.
    \item An evaluation of ViG on three different EO datasets, including a large-scale benchmark, and performance comparisons with a well-known vision transformer and CNN-based architecture.
\end{itemize}

The paper is structured as follows. Section~\ref{sec:related_work} introduces the related work. Section~\ref{sec:methodology} and Section~\ref{sec:experiments} introduce the problem's statement, the proposed methodology and the experimental results. Finally, Section~\ref{sec:conclusion} concludes the paper.

\section{Related Work}
\label{sec:related_work}
The ability to characterize land at large scale to identify and monitor resources (both natural and man-made) is known in remote sensing literature as the land cover and land use classification problem. Over the years, several different approaches were proposed by researchers, starting from the development of ad-hoc spectral indexes to automate the recognition of vegetation~\cite{ndvi_vegetation}, water-bodies~\cite{ndwi_water_segmentation} and burned areas~\cite{bais2}. Domain experts leveraged the characteristics of the response of different spectral bandwidths to ground elements to formulate indexes such as NDVI and NDWI, able to monitor the state of vegetation, water, and glaciers~\cite{ndvi_vegetation_regrowth, ndwi_glacier}. Consequently, such spectral imaging transformations defines mathematical transformations with several bandwidths, generating a single scalar value for each pixel. The output of a spectral index usually requires the evaluation of a field's expert to differentiate between the different classes (e.g., vegetation, water).
Moreover, such methodologies are highly sensible to noise, require ad-hoc calibration of thresholds and may vary depending on morphological, phenological features and biomes~\cite{different_thresholds}, and are not directly inferred from data.

Over the years, several approaches based on data inference were proposed, both unsupervised and supervised. \cite{isodata_kmeans_landcover} evaluated different clustering algorithms for land cover classification based on remote sensing acquisitions, whereas~\cite{fuzzyclustering_landcover_classification} proposed fuzzy clustering algorithm for land cover classification. Similarly,~\cite{hybrid_clustering} formulated a hybrid methodology to perform land use classification based on ISODATA algorithm.
Nonetheless, the research community also investigated the application of supervised approaches, such as random forests~\cite{random_forest_landcover} and SVMs~\cite{landcover_svm}.

As the data availability grew, as well as the presence of large-scale benchmark datasets~\cite{bigearthnet2, firerisk_dataset} and better performances achieved by neural networks, domain experts started adapting computer vision models to the aerial imagery context, proving superior performances. Purely convolutional models such as ResNet~\cite{resnet} demonstrated state-of-the-art performances in several different tasks, including land cover classification~\cite{landcover_imagenet}. More recently, the research community started investigating the application of transformer-based models in the EO field, confirming better performances than CNNs in case of high data availability~\cite{vit}.
Few applications of vision GNNs are present in literature, limited to small datasets~\cite{DING2022246, gnn_road_detection}. To the best of our knowledge, this paper introduces the first evaluation of a vision GNN, namely ViG~\cite{vig}, to a large-scale remote sensing multilabel benchmark and its adaptation to the EO domain.

\section{Methodology}
In this section, the problem of land cover classification is introduced, and a brief explanation of the ViG architecture is given. The model's configuration is explained in detail, with a paragraph describing the changes made to the original ViG model to adapt it to the remote sensing domain in case low-resolution images are considered.

\label{sec:methodology}
\subsection{Problem statement}
Given a dataset of satellite acquisitions, with each image being of size $H \times W \times C$, where $H$ is the height of the image, $W$ is its width, and $C$ is the number of channels ($C = 3$ for the RGB case, 12 for Sentinel-2 L2A imagery), the task of land cover classification automatically attributes to each input image $x_i$ a single class or multiple class labels, whether the problem is a multilabel classification problem or not.

\textbf{Multiclass classification.} Given an input image $x_i$, a set of $N$ possible labels to which each input image may belong to, and a prediction model $f_\theta$, trained on a set of labeled images, the model's output is an array of $N$ probability values $\hat{y}_j\in [0,1]$. Each value $\hat{y}_j$ represents the probability of the input image $x_i$ to exclusively belong to the j-th class. The probabilities sum to 1. Consequently, each input image $x_i$ is associated with the class with the highest probability according to model $f_\theta$.

\textbf{Multilabel classification.} Given an input image $x_i$, a set of $N$ possible labels to which each input image may belong to, and a prediction model $f_\theta$, trained on a set of labeled images, the model's output is an array of $N$ probability values $\hat{y}_j \in [0,1]$. Each value $\hat{y}_j$ represents the probability of the input image $x_i$ to belong to the j-th class. The probabilities do not sum to 1. Each satellite acquisition $x_i$ is associated with all the classes for which the predicted probability $\hat{y}_j > t$, being $t$ a predefined threshold set to 0.5.

\subsection{ViG model}

\begin{figure*}
    \centering
    \begin{subfigure}[b]{0.45\textwidth}
        \centering
        \includegraphics[width=\textwidth]{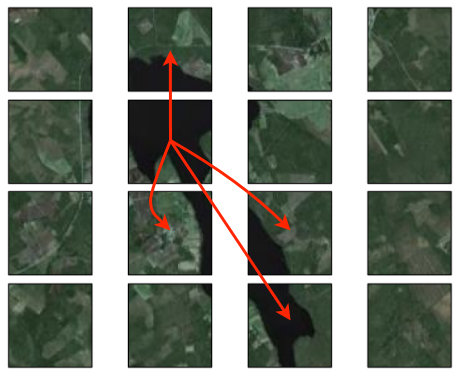}
        \caption{Lake acquisition.}
        \label{subfig:lake}
    \end{subfigure}
    \hfill
    \begin{subfigure}[b]{0.45\textwidth}
        \centering
        \includegraphics[width=\textwidth]{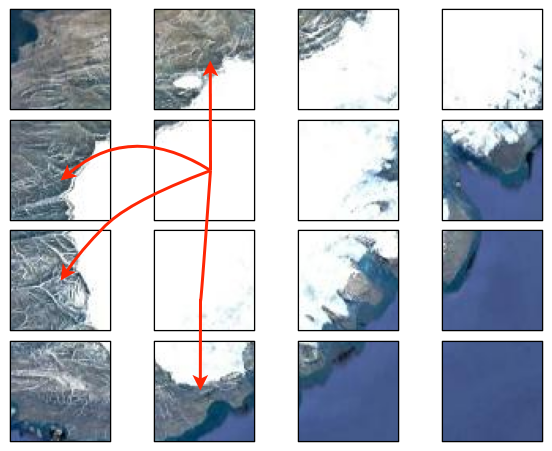}
        \caption{Snow acquisition.}
        \label{subfig:snow}
    \end{subfigure}
    \caption{Two samples extracted from RESISC45 and splitted into 16 patches belonging to class "Lake" (\ref{subfig:lake}) and class "Snowberg" (\ref{subfig:snow}) respectively. Red arrows represent possible dynamically-created edges with $K = 4$ in Grapher layer according to patch embeddings.}
    \label{fig:graph_wiring}
\end{figure*}

The main novelty introduced by Vision GNN architecture (ViG)~\cite{vig} is the decoupling of the computational graph from the image's regular grid structure and sequential structures adopted by CNN and transformer-based architectures. Instead, ViG splits the input image into several patches and projects them into a high-dimensional embedding space. The computational graph, i.e., the adjacency matrix adopted for the message passing in graph neural networks, is dynamically computed with a configurable number of neighbors based on similarity between embeddings. Each individual patch is treated as a node in a directed graph, with directed edges being created between the patch itself and the top-k most similar patches based on their embeddings. Once the graph has been defined, information is processed by a neural message passing mechanism, namely Grapher layer, consisting of max-relative graph convolution~\cite{graphconv} with a 2-layer MLP with ReLU nonlinearities. Grapher layer can be summarized as follows:
\begin{equation*}
    e = FC_b(\sigma(Grapher(FC_{nb}(x_i^h))))
\end{equation*}
\begin{equation*}
    z = FC_b(\sigma(FC_{nb}(e)))
\end{equation*}
where $FC_{nb}$ and $FC_b$ represent two fully connected layers without and with the bias term, respectively, whereas $\sigma$ is the ReLU nonlinearity. $z$ \text{and} $e$ represents Grapher's learned patch embeddings, before and after the projection of the feed-forward neural network (FFN) module, with $x_i^h$ being the input patch embedding.

ViG encoder consists of a sequence of Grapher modules followed by convolutional layers, each halving the original input resolution, i.e., after each grapher and convolutional module, the number of patches is reduced by a factor of 4.

An important characteristic to note is that ViG dynamically wires the graph depending on the input image and patch embeddings. Each patch is viewed as a node in the input graph, characterized by its patch embeddings. ViG automatically constructs a different graph at each Grapher layer (i.e., in our revised model, three different graphs are constructed, each with a different number of nodes due to the presence of downsampling modules) by computing for each patch the distance with respect to all the other patches. Given a predefined number of neighbors $K$, each patch is connected (in a directed way) to its K-nearest neighbor patches. More practically, given two input images $x_i$ and $x_j$ and given the same patch $h$ $x_i^h, x_j^h$, the neighbors of patch $x_i^h$ are likely to be in different positions compared to the neighbors of patch $x_j^h$. Patch position is not directly involved in edge creation: positional information is taken into account indirectly solely through patch embedding. An example is shown in Figure~\ref{fig:graph_wiring}.

The authors also presented a variant, namely PyramidViG, which builds pyramidal features and elaborates multiscale properties of the patches extracted, enhancing the model's performances~\cite{vig}. For our experimental evaluation, we implemented PyramidViG architecture, which will be referred to as ViG for brevity in Section~\ref{sec:experiments}.

Finally, image classification is performed with a simple classification head on top of ViG's encoder, composed of a single pooling layer and MLP classifier.

\subsection{ViG configuration}
\begin{figure*}[h!]
    \centering
    \includegraphics[width=.9\textwidth]{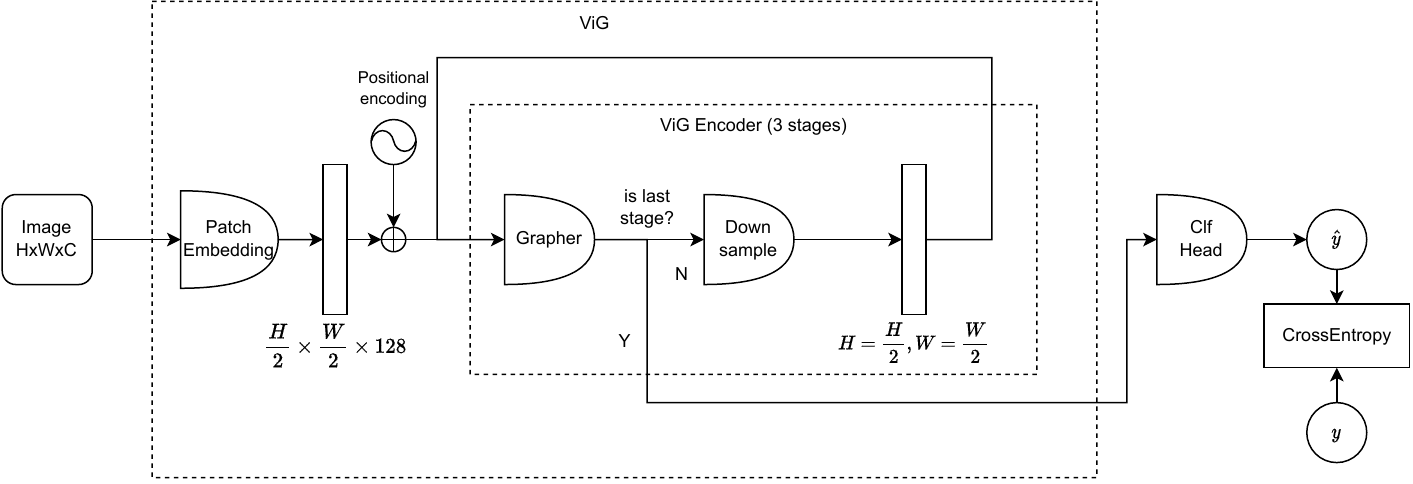}
    \caption{ViG's modified architecture. ViG Encoder block shows one single step of the encoder module, which is repeated three times. The downsample module is absent in the third encoder block. "Clf head" stands for "Classification head".}
    \label{fig:vig}
\end{figure*}

Due to the lower resolution of the input images in one of the benchmark datasets adopted in this work (120x120 pixels), we revised the ViG architecture to avoid the model collapsing onto a single patch. This is due to the fact that, after every Grapher layer, the height and width of the intermediate representation are halved, reducing at each step the number of patches by a factor of 4, leading to a very small number of patches in the last Grapher layer. To tackle such an issue, we modified the ViG's encoder structure to attain a similar number of learnable parameters compared to the original implementation and avoid the model's collapse in terms of the number of patches.
More in detail, we reduced the number of Grapher's stages from 4 to 3 with an embedding space of dimension [128, 256, 512] for the three stages instead of [48, 96, 240, 384] in the original implementation (ViG-Tiny). Additionally, we increased the number of heads from 4 to 16 and kept the number of layers in graph creation at $K = 9$ for every Grapher layer. The resulting model's size is introduced in Section~\ref{sec:experiments} and is differentiated for each dataset due to the varying number of input channels and input resolution in multispectral imaging. A schema of the resulting ViG architecture is shown in Figure~\ref{fig:vig}.

\section{Experiments}
This section describes the experimental settings, the benchmark datasets used, and the chosen hyperparameters. Finally, the experimental results are introduced and the performances achieved by ViG architecture are compared against the competitors.

\label{sec:experiments}
\subsection{Dataset}
We evaluated ViG on three open land cover classification benchmarks: two multiclass classification datasets and one multilabel classification dataset. The first two are RESISC45~\cite{resisc45} and PatternNet~\cite{patternnet}, both consisting of RGB-only acquisitions with a resolution of 256x256 pixels. Both datasets were collected using Google Earth, with a total number of satellite images available of 31500 and 30400, respectively. The RGB datasets are perfectly balanced, consisting of 45 and 38 distinct classes, each characterized by 700 and 800 samples respectively. RESISC45 is characterized by a resolution ranging from 0.2 to 30m per pixel, whereas PatternNet's resolution ranges from 6 to 50cm per pixel. Acquisitions in the RESISC45 dataset span over 100 countries with high variability in image conditions in terms of weather, morphological features of the terrain, and lighting.
The latter dataset is a large-scale multilabel image scene classification dataset, namely BigEarthNet~\cite{bigearthnet1, bigearthnet2}, composed of 590326 Sentinel-2 L2A and Sentinel-1 acquisitions (VV and VH polarizations), with a total number of 14 channels available. For our experiments, we limited the analyses on Sentinel-2's 12 bands. The highest resolution channels in BigEarthNet are of size 120x120 pixels, with a spatial resolution ranging from 60m to 10m per pixel, depending on the spectral band. Lower resolution acquisitions were upsampled with bilinear interpolation to match the highest resolution available, obtaining all input images of resolution 120x120 pixels.
The BigEarthNet dataset provides two highly imbalanced sets of labels: an easier task consisting of 19 labels and a harder one made of 43 different classes. We adopted the latter set of labels, considering only the higher-difficulty task.
All the datasets were split into train, validation, and test folds. For BigEarthNet, we adopted the split used by the original authors. For RESISC45, we chose the same split as in~\cite{resisc45split}, whereas for PatternNet, we used a 70/15/15 random non-overlapped split.
Table~\ref{tab:datasets} summarizes the different datasets' features.

Due to BigEarthNet's class imbalance, we report for each experiment the micro-averaged F1-Score, recall, precision, and accuracy. Instead, macro-average scores are reported for RESISC45 and PatternNet datasets because of their class-balanced distributions.

\begin{table*}[h!]
    \centering
    \caption{Split sizes (in terms of number of satellite acquisitions), number of distinct labels and other features of BigEarthNet, RESISC45, and PatternNet datasets.}
    \label{tab:datasets}
    \begin{tabular}{l|l|l|l|l|l|l|l|l|l|l}
        \toprule
        Dataset & Train & Validation & Test & \# labels & Resolution (pixels) & Resolution (m) & Bands & Coverage & Multilabel & Balanced \\
        \midrule
        BigEarthNet & 269695 & 123723 & 125866 & 43 & 120x120 & 10-60m & 12 (S2) & Europe & \cmark & \xmark \\
        RESISC45 & 18900 & 6300 & 6300 & 45 & 256x256 & 0.2-30m & 3 (RGB) & \begin{tabular}[c]{@{}l@{}}100+\\ countries\end{tabular} & \xmark & \cmark \\
        PatternNet & 21280 & 4560 & 4560 & 38 & 256x256 & 6-50cm & 3 (RGB) & Unknown & \xmark & \cmark \\
        \bottomrule
    \end{tabular}
\end{table*}

\subsection{Hardware, Code, and Reproducibility}
All the experiments were conducted on a workstation with one RTX A6000 48GB, 128GB of RAM, and Intel Core i9-10980XE 3.00GHz. All training procedures were run 3 times with 3 different seeds, reporting mean and standard deviation. Code to reproduce the experiments is written in Python 3.10, PyTorch~\cite{pytorch} and PyTorch Geometric~\cite{pyg} and is publicly available at~\url{https://github.com/lccol/vig-eo}.

\subsection{Baseline models}
We compared ViG performances with two other models: ResNet and ViT. In particular, we chose the ViG-Tiny model, having 7.1M trainable parameters in the original paper's formulation. Similarly, we chose ResNet18 and ViT-Tiny models for comparisons (11M and 5.5M, respectively) due to the larger sizes of ResNet34, ViT-Small, and ViT-Base (21.7M, 22M, and 86M) compared to ViG-Tiny. A comparison of the model sizes with respect to the datasets is shown in Table~\ref{tab:model_size}. The difference in number of trainable parameters for ViG-Tiny between BigEarthNet and RESISC45 is attributed to (i) the different number of input channels (12 vs. 3) and (ii) the higher input resolution (in pixels) for the latter two datasets in Table~\ref{tab:model_size}, leading to a significantly higher number of parameters assigned to the learnable positional encoding vector.

\begin{table}[h!]
    \centering
    \caption{Models' sizes in terms of number of trainable parameters.}
    \label{tab:model_size}
    \begin{tabular}{l|l|l|l}
    Dataset & ViT-Tiny & ViG-Tiny & ResNet18 \\ 
     \toprule
    BigEarthNet & 5.54M & 6.98M & 11.22M \\ 
    RESISC45 & 5.58M & 8.60M & 11.20M \\
    PatternNet & 5.58M & 8.60M & 11.20M \\
    \bottomrule
    \end{tabular}
\end{table}

\subsection{Training and loss functions}
We chose AdamW optimizer with a learning rate of $10^{-4}$ for all training procedures with weight decay ranging from $10^{-2}$ and $10^{-4}$ for at most 100 epochs in all datasets and models. Additionally, we adopted an early stop mechanism by monitoring validation loss with a patience of 10 epochs, reducing the learning rate with a factor 10 on plateau, using a patience of 5 epochs. The training was performed by minimizing the cross entropy loss in all the configurations. 
Training hyperparameters are summarized in Table~\ref{tab:train_hyperparams}.

\begin{table}[h]
    \centering
    \caption{Training hyperparameters.}
    \label{tab:train_hyperparams}
    \begin{tabular}{l|c|c|c}
    \toprule
     & ViG-Tiny & ViT-Tiny & ResNet18 \\
     \midrule
    Max epochs & \multicolumn{3}{c}{100} \\
    Optimizer & \multicolumn{3}{c}{AdamW} \\
    Learning rate & \multicolumn{3}{c}{$10^{-4}$} \\
    Weight decay & $10^{-4}$ & \multicolumn{2}{c}{$10^{-2}$} \\
    Early Stopping patience & \multicolumn{3}{c}{10 epochs} \\
    Early Stopping tolerance & \multicolumn{3}{c}{$10^{-3}$} \\
    LR scheduler & \multicolumn{3}{c}{Reduction on plateau} \\
    LR scheduler factor & \multicolumn{3}{c}{10} \\
    Batch size & \multicolumn{3}{c}{BigEarthNet: 256, Others: 16} \\
    \bottomrule
    \end{tabular}
\end{table}

\subsection{Experimental results and comparisons}
In this section, we compared to results obtained by the 3 considered models in the 3 datasets, in both multiclass classification (2 datasets) and multilabel classification (1 dataset) context. Due to the perfect class balance and high-class imbalance between the two cases, we evaluated macro-averaged metrics in the first case and micro-averaged scores in the latter.

\begin{table*}[h!]
\centering
\caption{Macro F1, Precision and Recall scores on RESISC45 and PatternNet datasets. Max and Min scores refer to the maximum and minimum per-class metrics. T stands for Tiny, Rec and Prec stand for Recall and Precision, respectively.}
\label{tab:resisc_patternnet_scores}
\setlength{\tabcolsep}{4pt}
\begin{tabular}{lllllllllll}
\toprule
         &     &         F1 &  Precision &     Recall & MaxF1 & MinF1 & MaxPrec & MinPrec & MaxRec & MinRec \\
Dataset & Model &                 &                 &                 &                  &                  &                         &                         &                      &                      \\
\midrule
PatternNet & ResNet18 &  99.68$\pm$0.02 &  99.69$\pm$0.02 &  99.69$\pm$0.02 &    100.0$\pm$0.0 &   97.89$\pm$0.27 &           100.0$\pm$0.0 &           \textbf{98.12$\pm$0.0} &        100.0$\pm$0.0 &       96.73$\pm$0.52 \\
         & ViG-T &   \textbf{99.7$\pm$0.07} &   \textbf{99.7$\pm$0.07} &   \textbf{99.7$\pm$0.07} &    100.0$\pm$0.0 &   \textbf{98.34$\pm$0.72} &           100.0$\pm$0.0 &          98.07$\pm$0.95 &        100.0$\pm$0.0 &       \textbf{97.62$\pm$0.48} \\
         & ViT-T &  99.06$\pm$0.01 &  99.07$\pm$0.01 &   99.06$\pm$0.0 &    100.0$\pm$0.0 &   93.67$\pm$0.87 &           100.0$\pm$0.0 &           94.82$\pm$0.4 &        100.0$\pm$0.0 &       92.56$\pm$2.06 \\
         \midrule
RESISC45 & ResNet18 &  83.69$\pm$0.18 &  83.86$\pm$0.19 &  83.83$\pm$0.13 &   98.03$\pm$0.32 &   56.35$\pm$1.05 &          97.18$\pm$0.73 &          57.35$\pm$2.38 &        98.9$\pm$0.38 &       55.48$\pm$2.18 \\
         & ViG-T &  \textbf{86.34$\pm$0.46} &  \textbf{86.49$\pm$0.45} &  \textbf{86.37$\pm$0.42} &   \textbf{99.37$\pm$0.44} &   \textbf{68.99$\pm$2.17} &          \textbf{99.75$\pm$0.44} &          \textbf{68.99$\pm$3.58} &       \textbf{99.34$\pm$0.44} &       \textbf{69.05$\pm$1.49} \\
         & ViT-T &  76.07$\pm$0.81 &   76.1$\pm$0.82 &  76.28$\pm$0.75 &   97.51$\pm$0.57 &    51.55$\pm$1.4 &          96.08$\pm$0.43 &          55.35$\pm$0.69 &       99.12$\pm$0.87 &       47.14$\pm$1.89 \\
\bottomrule
\end{tabular}
\end{table*}

\begin{table}[h!]
\centering
\caption{Micro average on F1, Precision and Recall scores on BigEarthNet dataset. T stands for Tiny.}
\label{tab:bigearthnet_scores}
\begin{tabular}{lllll}
\toprule
            &     &         F1 &  Precision &     Recall \\
Dataset & Model &                 &                 &                 \\
\midrule
BigEarthNet & ResNet18 &  75.11$\pm$0.56 &  80.74$\pm$0.66 &  70.23$\pm$1.45 \\
            & ViG-T &  \textbf{77.96$\pm$0.06} &  \textbf{82.78$\pm$0.42} &  73.68$\pm$0.36 \\
            & ViT-T &  77.75$\pm$0.07 &  82.28$\pm$0.21 &  \textbf{73.69$\pm$0.29} \\
\bottomrule
\end{tabular}
\end{table}

\subsubsection{Multiclass classification}
\label{sec:multiclass_exp}

The multiclass classification problem considered in this paper represents the easier task with mid-sized datasets being analyzed. Macro-average F1, precision, and recall scores are shown in Table~\ref{tab:resisc_patternnet_scores}. Moreover, we computed the maximum and minimum average per-class metric for each model. 

Evaluating results achieved on PatterNet, we can observe that the image classification task being considered is of lower difficulty compared to RESISC45. All models achieved scores higher than 99.0, with ViT being the worst among the three. ResNet's and ViG's F1, precision, and recall scores are similar, with ViT having metrics lower than 0.63 compared to the competitors. Instead, when evaluating the minimum per-class case, ViG achieved a minimum recall of 97.62, 5.06 percentage points (pp) higher than ViT and 0.89pp higher than ResNet. The vision GNN model achieves the best score across all the considered metrics, excluding minimum precision, in which ResNet outperformed ViG.

Experimental results instead demonstrate a higher complexity in solving the classification task proposed in RESISC45, obtaining a maximum average F1 score of 86.34 by ViG model, higher than 10.27pp and 2.65pp compared ViT and ResNet, respectively. The difference in performances gets amplified considering the lowest per-class metrics, in which ViG overperforms ViT by 17.44pp, 13.64pp, and 21.91pp in F1, precision, and recall, respectively.

Overall, learning image representations on a graph-based structure demonstrated state-of-the-art performances in land cover classification in middle-sized datasets, confirming the main limit of transformer-based architectures, requiring a great amount of data to learn better image representation.

\subsubsection{Multilabel classification}
Experiments on BigEarthNet dataset confirm the learning capabilities of ViG architecture in the context of multispectral imaging and remote sensing (12 input channels with RGB, infrared and ultra blue information). In this context, considering all the three micro-averaged metrics, ViG demonstrated better performances, followed by ViT (+0.21pp in F1) and ResNet being the worst (+2.85pp), despite having a lower number of parameters compared to the experiments in Section~\ref{sec:multiclass_exp}. This is caused by the significantly fewer number of patches and consequently fewer number of parameters associated with the positional encoding tensor.

\section{Conclusions}
\label{sec:conclusion}
In this paper, we assessed the performances of ViG image classification model in both multiclass classification and multilabel classification problems, demonstrating state-of-the-art performances in solving the problem of land cover classification. The considered architecture always surpassed ResNet performances with a lower number of parameters and achieved a superior F1 score (up to +10 percentage points) compared ViT on a medium-sized dataset characterized by high variability, thus demonstrating better learning capabilities and the necessity of a lower amount of training data compared to vision transformers. We release the code to the research community to foster research in the field of multispectral imaging, computer vision, Earth Observation, and graph neural networks.

In future developments, we plan to develop ad-hoc GNNs specifically for the Earth Observation domain, focusing on multimodal data (e.g., learning joint representations between SAR data, RGB, infrared and ultrablue bands) and extend our analyses to object detection in the remote sensing domain.

\section*{Acknowledgment}
The authors thank SmartData@PoliTO center for providing the computational resources.


\bibliographystyle{IEEEtran}
\bibliography{IEEEabrv,bibliography}

\end{document}